\documentclass[journal]{IEEEtran}
\usepackage{cite}
\usepackage{amsmath,amssymb,amsfonts}
\usepackage{algorithmic}
\usepackage{graphicx}
\usepackage{textcomp}

\usepackage{lineno,hyperref}
\usepackage{color}
\usepackage{subcaption}
\usepackage{multirow}
\usepackage{algorithm}

\begin{document}

\title{Spot The Odd One Out: Regularized Complete Cycle Consistent Anomaly Detector GAN}

\author{
    \IEEEauthorblockN{Zahra Dehghanian\IEEEauthorrefmark{1}, Saeed Saravani\IEEEauthorrefmark{1}, Maryam Amirmazlaghani\IEEEauthorrefmark{1}, Mohammad Rahmati\IEEEauthorrefmark{1}}\\
    \IEEEauthorblockA{\IEEEauthorrefmark{1}Department of Computer Engineering, Amirkabir University of Technology, Tehran, Iran\\
    Emails: \{z.dehghanian, s.saravani, mazlaghani, rahmati\}@aut.ac.ir}
}

\maketitle

\begin{abstract}
This study presents an adversarial method for anomaly detection in real-world applications, leveraging the power of generative adversarial neural networks (GANs) through cycle consistency in reconstruction error. Previous methods suffer from the high variance between class-wise accuracy which leads to not being applicable for all types of anomalies. The proposed method named RCALAD tries to solve this problem by introducing a novel discriminator to the structure, which results in a more efficient training process. Additionally, RCALAD employs a supplementary distribution in the input space to steer reconstructions toward the normal data distribution, effectively separating anomalous samples from their reconstructions and facilitating more accurate anomaly detection. To further enhance the performance of the model, two novel anomaly scores are introduced. The proposed model has been thoroughly evaluated through extensive experiments on six various datasets, yielding results that demonstrate its superiority over existing state-of-the-art models. The code is readily available to the research community at \href{https://github.com/zahraDehghanian97/RCALAD}{Github}.
\end{abstract}


\section{Introduction}\label{sec:Introduction}
	Discovering dissimilar instances and rare patterns is one of the
 most essential tasks in real-world data. Anomaly detection is the process of finding such samples, which are known as anomalies \cite{1yao2017anomaly}. Anomalies are an important aspect of any dataset and play an important role in a wide range of applications. For example, an irregular traffic pattern on a computer network could indicate hacking and data transmission to unauthorized places. Abnormalities in credit card transactions may reveal illicit economic activity \cite{2jiang2018credit}, or abnormalities in an MRI image may indicate the existence of a malignant tumor \cite{3dai2016distance}. Despite the existence of statistical and machine learning-based methods, designing effective models for detecting anomalies in complex, high-dimensional data spaces is still a major challenge \cite{4zenati2018adversarially}.
\\Generative adversarial networks (GANs) have shown remarkable performance in the field of anomaly detection by overcoming this challenge and modeling the distribution of high-dimensional, complex real-world data. In GAN, a generator network is contrasted with a discriminator network; the discriminator attempts to differentiate between the real data and the data produced by the generator network. The generator and  discriminator are trained simultaneously; the generator network G records the distribution of the data, and the discriminator D estimates the likelihood whether samples come from real data distribution or are being generated by G. The objective function of the generator G is to maximize the error probability of the network D. This structure leads to a two-player game like mini-max games \cite{5goodfellow2014generative}.
\\The ability of adversarial neural networks to represent natural images has previously been demonstrated \cite{6radford2015unsupervised,7creswell2018generative}, and their use in processing speech and text \cite{4zenati2018adversarially}, as well as medical images \cite{8schlegl2017unsupervised} is growing. This paper proposes an efficient method for anomaly detection that is based on generative adversarial networks. Similar to many learning-based algorithms, there are two main steps: training and testing. Like other adversarial frameworks, we train the generator and discriminator networks on normal data throughout the training phase so that both are updated in succession. In this case, the joint discriminator is used to make the training of the adversarial structure more stable. The encoder E is trained along with the discriminator and generator network to apply the inverse mapping of the input samples to the latent space. 

In order to minimize the distance between input space samples and their reconstruction and also minimize the distance between latent space samples (the inputs that are fed to the generator) and their reconstruction, we proposed a new joint discriminator named $D_{xxzz}$. In order to find an anomalous sample, the difference between input and its reconstruction is computed and samples with high differences can be determined as anomalies. Finding anomalous samples necessitates an appropriate distance between the input sample and its reconstruction. To achieve this aim, we also train our network on some samples that are produced by the proposed supplementary distribution $\sigma(\mathrm{x})$.In summary, this paper provides the following contributions:
                	\begin{itemize}
		            \item A new discriminator $D_{xxzz}$ in order to provide complete cycle consistency is proposed. 
		              \item A supplementary distribution $\sigma(\mathrm{x})$ is used in order to bias network towards normal data manifold.
		              \item Two new anomaly scores are introduced.
	                \end{itemize}
The structure of this paper is as follows: In the next section, we have a summary of previous works. Preliminaries are provided in Section 3. The proposed model is elaborated in Section 4. The experiments will be discussed in Section 5. We have a conclusion in Section 6, and the final section is devoted to future work.

\section{Related work}\label{sec:Related_work}
Anomaly detection, also known as novelty detection and outlier detection, has been widely studied, as reviewed in \cite{9kaur2016survey,10zimek2012survey,11pimentel2014review}. The previous methods used in this field are generally divided into two categories: methods based on representation learning and methods based on generative models. 

A representation learning method learns a mapping for the main characteristics of normal data. One-class support vector machine finds the marginal boundary around the normal data \cite{12scholkopf1999support}. The isolation forest method is one of the classic machine learning methods. In this method, the tree is built with randomly chosen features, and the anomaly score is the average distance to the root \cite{13ruff2018deep}. Deep support vector data description (DSVDD) finds a hypersphere to enclose the representation of normal samples \cite{14liang2017enhancing}. Liu and Gryllias constructed frequency domain features using cyclic spectral analysis and applied them in the support vector data description (SVDD) framework. This method has been proven robust against outliers and can achieve a high detection rate for detecting anomalies \cite{15liu2020semi}. In \cite{16golan2018deep}, researchers presented a new approach to identify imagery anomalies by training the model on normal images altered by geometric transformation. In this model, the classifier calculates the anomaly score using softmax statistics.

Usually, generative models attempt to learn the reconstruction of the data and use this reconstruction to identify anomalous samples \cite{17yang2020regularized}. For instance, auto-encoders model the normal data distribution, and the reconstruction error is used as the anomaly score \cite{18nguyen2019anomaly,19pidhorskyi2018generative}. Deep structured energy-based models (DSEBMs) learn an energy-based model and map each sample to an energy score \cite{20zhai2016deep}. Deep autoencoding Gaussian mixture model (DAGMM) estimates a mixed Gaussian distribution by using an encoder for normal samples \cite{21schlegl2019f}. A recent line  of work on anomaly detection has focused on adversarial neural networks. For example, this structure has been used to identify anomalies in medical images \cite{8schlegl2017unsupervised}. In this work, the inverse mapping into the latent space was performed using the recursive backpropagation mechanism. In \cite{22zenati2018efficient}, a continuation of the prior work, the mapping to the latent space was performed by the encoder in order to reduce computational complexity. In \cite{23dumoulin2016adversarially}, the proposed model was based on bidirectional GAN (BiGAN). The encoder was also responsible for mapping the input data space to the latent space. Unlike the standard GAN structure, where the discriminator takes only the real image and the generated image of the generator network as input, the representation of these images in the latent space was also considered as input to the discriminator network. 

In the article \cite{4zenati2018adversarially}, anomaly detection was performed by adding a new discriminator in latent space to the adversarial structure to stabilize the training process. {In \cite{li2019mad} a gan-based model is utilized to detect anomalies in time series. In the context of Mad-gan, the generator is trained on normal time series to learn the underlying patterns and dependencies. It then generates synthetic samples that resemble the normal data distribution. The discriminator is simultaneously trained to differentiate between real and synthetic samples. DCT-GAN proposes a novel approach for detecting anomalies in time series data \cite{li2021dct}. The method combines dilated convolutions and transformer architecture within a Generative Adversarial Network (GAN) framework. The dilated convolutions capture long-range dependencies in the time series, while the transformer module learns the temporal relationships between different time steps. The generator network generates synthetic samples, and the discriminator network distinguishes between real and fake samples. By training the GAN on normal time series data, it learns to generate realistic normal samples. Anomalies can be detected by measuring the discrepancy between real and generated samples using a reconstruction loss. Experimental results demonstrate that DCT-GAN outperforms existing methods in terms of anomaly detection accuracy on various benchmark datasets.

Many existing GAN-based anomaly detection algorithms have weak robustness, in \cite{li2023m3gan} the authors introduce a technique called M3gan, which combines a masking strategy with a mutable filter to diversify the data and improve the robustness of the model. The M3gan approach starts by dividing the multidimensional data into smaller subsets called cells. Each cell is then masked by randomly selecting a subset of its dimensions and replacing their values with noise. This masking strategy helps in preserving the privacy of sensitive information while still allowing for accurate anomaly detection. } 
The deep convolutional autoencoder model is a classic autoencoder model in which the encoder and decoder have a convolutional structure. The anomaly score in this model is the 2-norm of reconstruction errors \cite{24li2017alice}. {In \cite{ruff2019deep} a novel deep learning framework for semi-supervised anomaly detection called deep SAD (semi-supervised anomaly detection) has been proposed. The proposed method, called Deep SAD, combines the benefits of both supervised and unsupervised approaches by leveraging a small amount of labeled normal data and a large amount of unlabeled data. Deep SAD utilizes an autoencoder-based architecture to learn a low-dimensional representation of the input data and employs a novel loss function that encourages the model to assign low reconstruction errors to normal instances while penalizing anomalies. }

In this section, we reviewed and categorized the various methods used to clarify abnormal data. In section 3 we will deep into fundamental basis required to deal with proposed model.   

 \section{Preliminaries}
In this section, we briefly elaborate the overall idea of a generative adversarial network and then concentrate on the evolution of GAN-based anomaly detection algorithms. The generative adversarial network was first proposed in 2014 by Goodfellow et al \cite{5goodfellow2014generative}. The generator and the discriminator are trained on an M set of $\left\{\chi^{(i)}\right\}_{i=1}^{M}$ unlabeled samples. The generator maps selected samples from the latent space z to the input data space. The discriminator attempts to distinguish between the real data $x^{(i)}$ and the data produced by the generator (G). The generator G imitates the input data distribution, whereas the discriminator differentiates between real samples and generator data. In the training phase, the generator G and the discriminator D are alternatively optimized using a stochastic gradient descent approach.

$q(x)$ is the distribution of the input data, and $p(z)$ is the distribution of the generator in the latent space. GAN network training is done by finding a discriminator and a generator that can solve the saddle point problem as  $\min _{G} \max _{D} V_{G A N}(D, G)$  and the definition of the $V_{G A N}(D, G)$ function is defined as: 
\begin{equation}
V_{G A N}=E_{x \sim q(x)}[\log (D(x))]+E_{z \sim p(z)}[\log (1-D(G(z)))]
\end{equation}

Solving this problem concludes that the generator distribution is equal to the true data distribution. It has been proved in \cite{5goodfellow2014generative} that the global optimal discriminator will be obtained if and only if $p_G (x)=q(x)$. By $p_G$, we mean the distribution learned by the generator.
Adversarially learned inference (ALI) \cite{23dumoulin2016adversarially} attempts to obtain the inverse mapping between input data space and latent space by modeling the joint distribution of encoder as $q(x,z)=q(x)e(z|x)$ and the distribution of generator as $p(x,z)=p(z)p(x|z)$ using the encoder E. Here, $e(z|x)$ is learned by the encoder. The objective function of the ALI model is as follows:

\begin{equation}
\resizebox{\columnwidth}{!}{%
$\begin{gathered}
\min _{G, E} \max _{D} V_{A L I}=\\E_{q(x, z)}[\log D(x, E(x))]+E_{p(x,z)}[\log (1-D(G(z), z)]]
\end{gathered}$
}
\end{equation}
where D represents the discriminator, taking $x$ and $z$ as input, and its output value specifies probability of origination of the current inputs from the $q(x,z)$  distribution. Encoder, generator, and discriminator are in their optimal state only if $q(x,z)=p(x,z)$. This has been proved in \cite{23dumoulin2016adversarially}. 

Although $p$ and $q$ distributions are apparent, in practice and during model training, they are not necessarily converging to the optimal point. This issue was attributed to the problem of cycle consistency, which was defined as $\mathrm{G}(\mathrm{E}(\mathrm{x})) \approx \mathrm{\hat{x}}$ in\cite{24li2017alice}. A new framework called ALICE was proposed to solve the above problem by adding the discriminator $\mathrm{D}_{\mathrm{xx}}$ to the ALI network structure \cite{24li2017alice}. The objective function of this model is as follows:
\begin{equation}
\resizebox{\columnwidth}{!}{%
$\begin{gathered}
\min_{E, G} \max_{D_{x z}, D_{x x}} V_{A L I C E}=V_{A L I}+\\E_{x \sim q(x)}\left[\log D_{x x}(x, x)+\log (1-D_{x x}(x, G(E(x))))\right]
\end{gathered}$
}
\end{equation}
This work demonstrates that using a discriminator $\mathrm{D}_{\mathrm{xx}}$ can achieve the best reconstruction for the input data \cite{26zong2018deep}. In Adversarially Learned Anomaly Detection (ALAD), a conditional distribution was applied to the baseline ALICE model with the inclusion of additional discriminator to stabilize the training process \cite{4zenati2018adversarially}. To deep into the detail, a discriminator $\mathrm{D}_{\mathrm{zz}}$ is added to the model to ensure the cycle consistency in the latent space, which tries to make the latent space variable and its reconstruction as analogous as possible. By assembling the block proposed in \cite{4zenati2018adversarially} in ALICE framework, the cost function of the ALAD model will finally be as follows:
\begin{equation}
\resizebox{\columnwidth}{!}{%
$\begin{gathered}
\min _{G, E} \max _{D_{x z}, D_{x x}, D_{z z},} V_{A L A D}=V_{A L I C E}+\\\mathbb{E}_{z \sim p(z)}\left[\log \left(D_{z z}(z, z)\right)+\right.
\mathbb{E}_{z \sim p(z)}\left[\log \left(1-D_{z z}(z, G(E(z)))\right)\right]
\end{gathered}$
}
\end{equation}

In \cite{4zenati2018adversarially}, it is claimed that the training model will be stabilized by adding Lipschitz constraints to the discriminators of the GAN model. Moreover, it is shown in practice that with spectral normalization of the weight parameters, the network's performance will be improved \cite{4zenati2018adversarially}.

Although the idea of ALAD helps stabilize the cycle, the latent and input space variables are scrutinized in two independent spaces, and the inherent dependence between the variables is ignored. More precisely, the x and its reconstruction are investigated in a separate process from the z and its corresponding reconstruction, while the reconstruction processes of these two pairs of data is along with each other and affects each other directly. To model this reliance, we define a complete cycle and a new discriminator that employs the information contained within the complete cycle.

Another problem with ALAD is that it doesn't take into account the need for weak reconstruction for anomalous samples. In fact, in all of the previous models, it was assumed that if the model were trained with normal data, it would always have a poor reconstruction for anomalous data, even though there is no way to force the model to make weak reconstructions of anomalous samples. The proposed RCALAD model attempts to address this issue by employing the supplementary distribution $\sigma(\mathrm{x})$, which biases all reconstructions toward the normal data distribution, and the model attempts to lower the anomaly score for normal input while increasing it for anomalous input.
\section{Proposed Model}
In this part, we describe our novel adversarial framework for anomaly detection that overcomes the aforementioned drawbacks. The problem of a lack of complete cycle consistency and our corresponding solution will be described first. Then, the necessity of weak reconstruction will be addressed. At the end of this section, two new anomaly scores based on the proposed model are introduced.
\subsection{Complete Cycle Consistency}
As mentioned before, in the previous models, the cycle consistency of the input data and latent space variables is examined in two independent procedures. This implies that the reconstruction of the latent space variable $\hat{z}$ and input data $\hat{x}$ are processed separately. {In other words, the previous works tried to resemble $x$ and its reconstruction and $z$ and its reconstruction in two separate cycles. The inverse mapping of $x$ was neglected in the z-cycle, and the direct mapping of $z$ was also ignored in the x-cycle. To tackle this problem, we introduce a new complete cycle that simultaneously tries to learn joint distribution, which utilizes neglected parameters in a complete cycle that contain rich information.}  It should be noted that here the variable $z$ is a sample of the Gaussian distribution given as an input to the generator, and it is not related to the input mapping in the latent space.

The Complete Cycle Consistency issue (CCC) declares that for each variable $x$ in the input space if the encoder first estimates the inverse mapping to the latent space, which equals $E(x)=z_{x}$ and the obtained representation is entered into the generator to generate the network reconstruction from the input variable $G\left(z_{x}\right)=G(E(x))=\hat{x}$ and the this reconstruction is given to the encoder network in order to calculate the reconstruction in the latent space, that is, $E(\hat{x})=E\left(G\left(z_{x}\right)\right)=\hat{z}_{\hat{x}}$, it is logically expected from any reconstruction-based network that the two variables $x$ and $\hat{x}$  as well as the two variables $\hat{z}_{\hat{x}}$  and ${z}_{x}$ have the least possible difference. That is, the CCC issue is defined in such a way that, in any reconstruction-based model, for each input sample and its mapping in the latent space, the network reconstruction for both variables should have a minimum error and maximum similarity.

\begin{figure}[t]
  \centering
  \includegraphics[width=1\columnwidth]{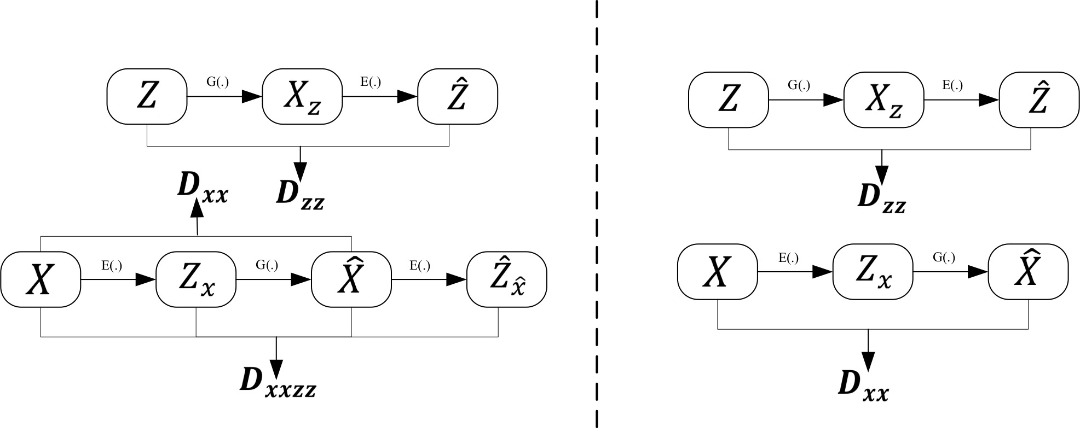}

   \caption{ {The information of complete cycle consistency in the proposed model (left side) and using the variables of input data space and latent space in the cycle consistency of the ALAD network (right side).}}
   \label{fig10}
\end{figure}

Without using the CCC, the similarity between the input data and its reconstruction, as well as the similarity between $z$  and its reconstruction, were examined independently and in two separate cycles. It was assumed that they are independent, but we know that these two cycles are entirely dependent on each other, and the assumption of independence is not valid in these two issues. To solve this problem, we proposed to model the dependency by examining the CCC variables in the new discriminator $D_{xxzz}$ and using the information flow in this chain to improve network training for anomaly detection in the best possible way. The difference between the input of $D_{xxzz}$ and the input of $D_{zz}$ used in the ALAD model is represented in Figure \ref{fig10}. 

As can be seen in Figure \ref{fig10}, the ALAD model does not use the information of a complete cycle. In order to use the available joint information in a complete cycle, a new variable called $\hat{z}_{\hat{x}}$ is introduced. To calculate this variable, the inverse mapping of the input data x is applied to the generator, and the resulting inverse mapping is calculated again using the decoder. Hence, complete cycle consistency will be provided in this model.

In order to ensure the condition of complete cycle consistency, the new $D_{xxzz}$ discriminator is used with the joint input. It is noteworthy that the effectiveness of the joint discriminators has already been proven once in ALIGAN \cite{23dumoulin2016adversarially}. Actually, when adding the encoder to the GAN framework, two procedures can be scrutinized. The first one is adding an independent discriminator to train the encoder, and the second one is changing the discriminator input from a single input mode to a joint input mode. It is proven that using a joint discriminator obtains better results. According to the same idea, the input of the joint discriminator $D_{xxzz}$ extracts the most information for model training.

This discriminator uses the quadruple
$\left(x, x, z_{x}, z_{x}\right)$  as the real data and the quadruple of $\left(x, G(E(x)), z_{x}, E\left(G\left(z_{x}\right)\right)\right)$ as the fake data. This discriminator attempts to make the input $x$ and network reconstruction, as well as the inverse mapping of the input image in the latent space and its reconstruction by the encoder, as close as possible to each other so that a complete stable loop is provided and the model is trained and stabilized better.
\subsection{Constraint of Weak Reconstruction}
In reconstruction-based models, it has always been assumed that if the training and reconstruction of the normal data are properly done, the reconstruction of abnormal data will necessarily be weak and different from the input data. However, experiments indicate that it is not always the case, and sometimes the reconstructed anomalous sample is slightly similar to the input sample. Hence, it won't be easy to recognize it as an abnormal sample. In fact, in none of the previous models, there was not any obligation or control condition to bias the model toward producing poor reconstructions for anomalous samples.

\begin{figure}[t]
  \centering
  \includegraphics[width=1\columnwidth]{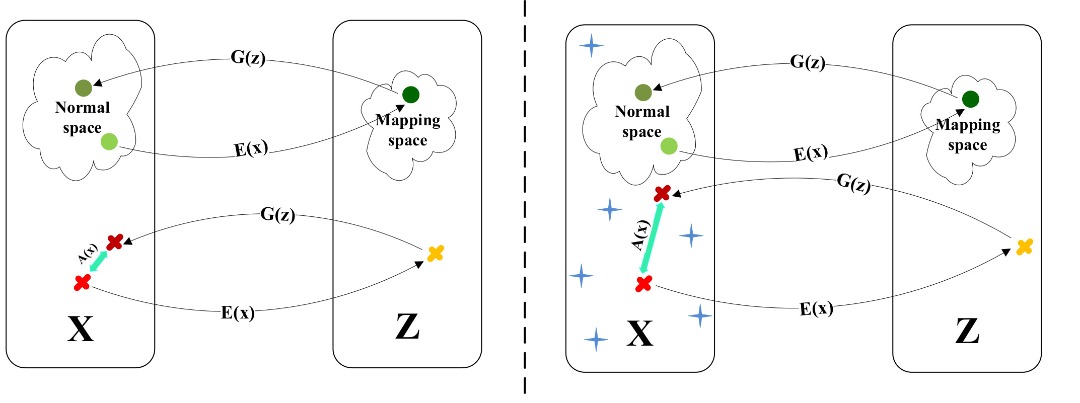}
   \caption{{Effect of the presence of supplementary distribution in the model training process. These figures indicate the trained models. On the left side, there is no supplementary distribution in the training phase, and on the right side, we had samples from supplementary distribution in training. In this figure, \textbf{X} represents the input data space and \textbf{Z} represents the latent space. Samples are mapped from the latent space to the input data space by the generator $G$, and the encoder $E$ reverses the mapping. Green circles show normal samples, red crosses represent abnormal samples, and blue stars represent samples generated by the supplementary distribution $\sigma(\mathrm{x})$. The turquoise-colored line shows the value of the abnormality score. As can be seen in Figure \ref{fig2}, if samples from the supplementary distribution $\sigma(\mathrm{x})$ are not present (on the left side of the figure), the abnormality score for the abnormal sample is lower than when these samples(blue stars) are used used (right side). In other words, when there is no supplementary distribution in the training process, after training, the model may reconstruct anomalous samples in a good way, which leads to a small difference between anomalous sample and its reconstruction, and as a result, we will have a low anomaly score, but on the right, the samples from supplementary distribution have biased the model towards the normal manifold. In this case, the model tries to reconstruct an anomalous sample near the normal data manifold, which is far from the anomalous data manifold that leads to a high anomaly score for abnormal data.}}
   \label{fig2}
\end{figure}

During the training phase, the encoder and generator are only trained on normal samples. As a result, the appropriate $z$ space for normal samples and the reconstruction of them is well modeled. However, because the model has not yet seen the remainder of the space, including abnormal examples, it may map it onto an unknown point in the latent space during the testing phase. In this case, there is no information about mapping anomalous data to its reconstruction. To solve this problem, the supplementary distribution called $\sigma(\mathrm{x})$ is used. Samples from this distribution will cover the input data space, which means that we will produce some extra noisy samples and force the model to generate reconstruction in the normal data manifold. {Here we teach the model to produce output in a normal data manifold by feeding it with a wider range of inputs. Through using this distribution, we have further training with unseen data, and as a result, we could have a generalization that will help model produce reconstructions in normal data manifold for a wider range of inputs compare to the absence of the supplementary distribution.} Through this extra training, the network learns to reconstruct the normal data class for a relatively more expansive range of inputs. So an appropriate distance is created between the anomalous sample and its reconstruction. This distance is regarded as an appropriate criterion for detecting abnormal samples. Figure \ref{fig2} illustrates the effect of using $\sigma(\mathrm{x})$ in training routine.
\subsection{RCALAD Model}
\begin{figure}[t]
  \centering
  \includegraphics[width=1\columnwidth]{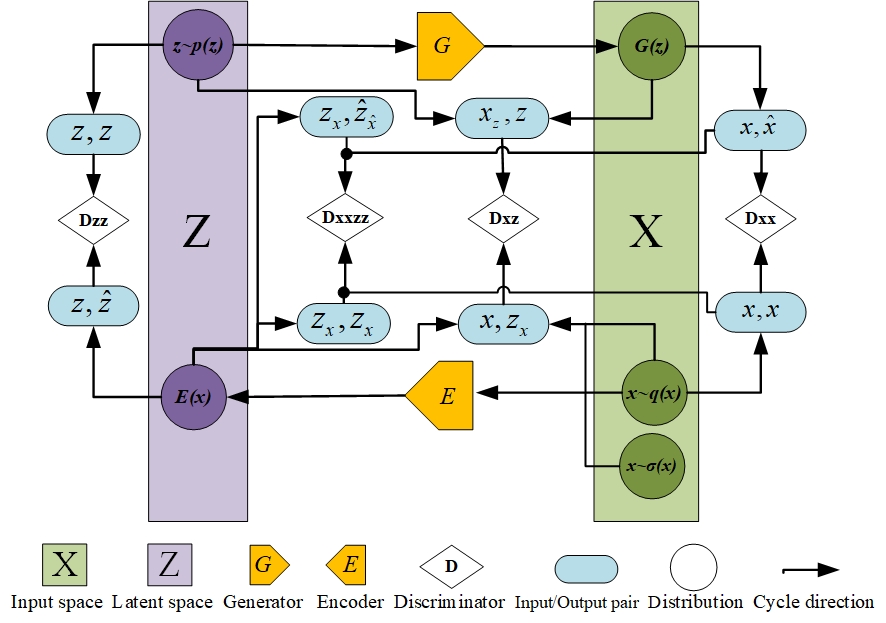}
   \caption{Overall structure of the RCALAD model.}
   \label{fig3}
\end{figure}

In this section, we will introduce our proposed model, RCALAD, by integrating both ideas, which means employing the new variable of $\hat{z}_{\hat{x}}$  in the $D_{xxzz}$ discriminator and the $\sigma(\mathrm{x})$ distribution and adding them to the basic model \cite{4zenati2018adversarially}. In this network,  we attempted to provide a comprehensive, practical, and compatible framework for all anomaly detection problems by resolving the aforementioned issues of the complete consistency cycle and the necessity of weak reconstruction. Figure \ref{fig3} depicts a schematic representation of the proposed model.
{As we can see in Figure \ref{fig3}, an encoder and the generator are trained in the standard structure of the adversarial neural network. The inverse mapping from the input data space to the latent space is obtained simply by using the encoder $E$ in the proposed structure  and the direct mapping from the latent space to the input data space is done by the generator $G$. Here $D_{xz}$ is a new name for a standard discriminator needed for adversarial training of $G$ and $E$. It is responsible for discriminating between two pairs: $(z,G(z)$ and $(x,E(x))$. Through this discrimination, the encoder $E$ learns the inverse mapping of input $x$, and also the generator $G$ learns the direct mapping from the latent space to the input data space. This discriminator determines whether an input variable pair is derived from the input data $x$ distribution and its corresponding point in the latent space or if it is generated by the generator $G$ and sampled from the latent space of $z$. In order to satisfy the cycle consistency condition in the input and latent space, $D_{xx}$ and $D_{zz}$ discriminators are used, so cycle consistency in both spaces will be modeled independently. Here, a joint discriminator called $D_{xxzz}$ is used to train both generator and encoder networks simultaneously. The $D_{xxzz}$ is introduced to use all the information in a complete cycle.} That is, in addition to examining both variables $x$ and $z$ and their reconstruction $\hat{x}$ and $\hat{z}$ in the corresponding space, their joint distribution is used in $D_{xxzz}$ during the process of detecting anomalous samples. By using $D_{xxzz}$ more information is available to determine whether the input data is anomalous or not. This network is responsible for determining between quadruple samples of $\left(x, x, z_{x}, z_{x}\right)$ and $\left(x, G(E(x)), z_{x}, E\left(G\left(z_{x}\right)\right)\right)$. In fact, the discriminator $D_{xxzz}$ tries to maximize the similarity between $x$ and its reconstruction $G(E(x))$, and it also attempts to make the mapping of the input image in the latent space $z_{x}$ and its reconstruction $E(G(z_{x}))$ as similar as feasible. To cover much more of the latent space, the $\sigma(\mathrm{x})$ block is added to this model. Using this distribution, new samples are made from the input data space and then mapped into the latent space of the normal data. Finally, the objective function of the proposed model is as follows:

\begin{equation}
\resizebox{\columnwidth}{!}{%
$\begin{gathered}
\min _{G,E} \max _{D_{x x z z},D_{x z},D_{x x},D_{z z}}
V_{R C A L A D}\left(D_{x x z z},D_{x z},D_{x x},D_{z z},E,G\right)=\\
V_{A L A D} +  E_{x \sim \sigma(x)}\left[\log \left(1-D_{x z}(x,E(x))\right)\right]+\\
E_{x \sim \mathrm{q}(\mathrm{x})}\left[\log D_{x x z z}(x,x,E(x),E(x))\right]+\\
E_{\boldsymbol{x} \sim q(x)}\left[1-\log D_{x x z z}(x,G(E(x)),E(x),E(G(E(x))))\right]
\end{gathered}$%
}
\end{equation}

\subsection{Anomaly Scores}
The main goal of this proposed model is to detect anomalies based on the accurate reconstruction of normal input data, whereas abnormal samples are reconstructed in a weak manner. One of the key elements in anomaly detection is the definition of the anomaly score for calculating the distance between the input sample and the reconstruction provided by the network \cite{4zenati2018adversarially}. Some of the anomaly scores that were used in previous models are as follows:
\begin{equation}
\begin{gathered}
A_{L_{1}}(x)=\|x-\hat{x}\|_{1}, \\
A_{L_{2}}(x)=\|x-\hat{x}\|_{2}, \\
A_{\text {Logits }}(x)=log \left(D_{x x}(x, \hat{x})\right),\\
A_{\text {Features }}(x)=\left\|f_{x x}(x, x)-f_{x x}(x, \hat{x})\right\|_{1}
\end{gathered}
\end{equation}
Here, logit means the raw output of the discriminators, while feature means the output of the layer preceding logit. In our proposed model, since the new discriminator $D_{xxzz}$ offers the ability of extracting additional information, it is crucial to specify the new anomaly scores in order to make use of this information. Therefore, two new anomaly scores are introduced.

The first anomaly score presented in this paper is called $A_{f m}(x)$. This score uses the $D_{xxzz}$ discriminator feature space to calculate distance between samples and their reconstruction. For this purpose, the output of the second-to-last layer is used as features. Our anomaly score is defined as follows:
\begin{equation}
\begin{gathered}
A_{fm}(x)=\left\|f_{x x z z}\left(x, x, z_{x}, z_{x}\right)-f_{x x z z}\left(x, \hat{x}, z_{x}, \hat{z}_{\hat{x}}\right)\right\|_{1}
\end{gathered}
\end{equation}
In this equation, $f(.)$ represents the activation function of the next to last layer in the $D_{xxzz}$ discriminator structure. The concept used in the definition of this score is using the confidence level of the discriminator on the quality of the reconstructions provided by the network. In other words, if reconstruction is performed well, the sample belongs to the trained normal data of the network. Thus, the higher value of this criterion means the greater difference in reconstructions and so higher possibility of input data's abnormality. The second point in this article is presented with the aim of maximizing the use of information in the model for anomaly detection. In this section, the $A_{all}$ criterion is defined. The score is the sum of the outputs of all discriminators, including $D_{xx}$, $D_{zz}$ and $D_{xxzz}$.

In fact, since all the discriminators in the proposed model are trained only based on the normal samples and the reconstruction for all the input data space is biased towards the normal data space, it is expected that the input image data and its reconstruction will look different, and the discriminators can easily identify these anomalous inputs. The mathematical expression of this criterion is given in the following equation:
\begin{equation}
\begin{gathered}
A_{all}(x)=\frac{1}{3}
(D_{x x z z}\left(x, \hat{x}, z_{x}, \hat{z}_{\hat{x}}\right)+D_{x x}(x, \hat{x})+D_{z z}\left(z_{x}, \hat{z}_{\hat{x}}\right))
\end{gathered}
\end{equation}
The criterion $A_{all}$ tries to utilize all discriminators' information. During the training phase, the discriminators learn to pay attention to the difference between the pairs of $(x,x)$ and $(x,\hat{x})$ as well as the pairs of $(z_{x},z_{x})$ and $(z_{x},\hat{z}_{\hat{x}})$. It means, the farther $\hat{x}$  from $x$ or $\hat{z}_{\hat{x}}$  from $z_{x}$, it will be easier for the discriminators to recognize the data origin. In the proposed model, by adding the distribution of $\sigma(\mathrm{x})$ and biasing all the reconstruction towards the normal data distribution, the reconstruction error for abnormal data is increased and the discriminators' output can be considered as a reliable criterion for abnormality detection. Finally, the recommended anomaly scores can be viewed according to the algorithm \ref{alg1}.


\begin{algorithm}
\caption{Process of calculating anomaly scores in Regularized Complete Adverarially Learned Anomaly Detection}
\label{alg1}
\begin{algorithmic}
\STATE {\bfseries Input:} $x \sim p_{x_{\text {Test }}}(x)$ , $E$, $G$, $D_{xx}$, $D_{zz}$, $D_{xxzz}$, $f_{xxzz}$ \%  $f_{xxzz}$ is the feature layer of $D_{xxzz}$
\STATE {\bfseries Output:} $A_{all}{(x)}$, $A_{fm}{(x)}$ \% $A$ is the anomaly score
\STATE procedure INFERENCE
\STATE $\, {z_{x}} \leftarrow {E}({x})$ \%Encode samples, Construct latent Embedding
\STATE $\, \widehat{{x}} \leftarrow {G}\left({z}_{x}\right)$ \%Reconstruct samples
\STATE $\, \hat{{z}}_{\hat{x}} \leftarrow {E}(\widehat{{x}})$ \%Reconstruct latent Embedding
\STATE $\, A_{fm}({x})\leftarrow\left\|{f}_{x x z z}\left({x}, {x}, z_{x}, z_{x}\right)-{f}_{x x}\left({x},\widehat{{x}}, {z}_{x}, \hat{{z}}_{\hat{x}}\right)\right\|_{1}$
\STATE $\, A_{all}({x}) \leftarrow \frac{1}{3}
(D_{x x z z}\left({x}, \widehat{{x}}, {z}_{x}, \hat{{z}}_{\hat{x}}\right)+ D_{x x}({x}, \hat{{x}})+ D_{z z}\left(z_{x}, \hat{{z}}_{\hat{x}}\right))$
\STATE \textbf{$\,$return} $A_{a l l}({x}), A_{f m}({x})$
\STATE \textbf{end} procedure
\end{algorithmic}
\end{algorithm}

\section{Experiments}

This section compares the proposed RCALAD model with prominent anomaly detection models. To test the models on a fair basis, the reported outcomes for all the implemented models are based on tabular data obtained from the average of ten runs, and for each class of image data, they are based on the average of three runs. The anomaly score used in tabular data is $A_{all}$ score and, for image data, it is $A_{fm}$ score. The reason for choosing these scores based on data type will be discussed later. Moreover, the ALAD model is implemented and the results of the best anomaly score $A_{Features}$ are reported. For other models, the available results are adopted from \cite{4zenati2018adversarially,17yang2020regularized} . {In testing phase, anomalous samples and normal samples are fed to the trained model. We expect that the model produce a high anomaly scores for anomalous samples and low anomaly scores for normal samples. We know that for KDDCup99, arrhythmia, thyroid, and musk datasets 20, 15, 2.5 and 3.2 percent of the data are anomalous samples, respectively, so we consider the samples with high anomaly scores as anomalous samples. For the CIFAR-10 and SVHN 90 percent of the data are anomalous.}
\subsection{Datasets}

In order to evaluate the performance of the proposed model and scrutinize its efficiency from different viewpoints, various datasets with diverse characteristics are used. The proposed method is tested on the available image and tabular datasets. For tabular datasets, four datasets, including KDDCup99\cite{misc_kdd_cup_1999_data_130}, arrhythmia\cite{misc_arrhythmia_5}, thyroid\cite{misc_thyroid_disease_102}, and musk \cite{misc_musk_(version_1)_74}, are used. The KDDCup99 dataset contains nearly 5000000 samples from 41 dimensions, 34 of which are continuous and 7 of which are categorical. We then utilized one-hot representation to encode categorical characteristics, yielding a total of 121 features following this encoding. Arrhythmia is a medical collection related to cardiac arrhythmia with 16 classes. {This database contains 274 attributes and it has 490 samples. We applied our methods on raw data.} Also, thyroid is a three-class dataset related to thyroid disease. {The “hyperfunction” class, consisting of 2.5\% of the dataset set, is treated as anomaly.It contains 3772 samples and each sample has 6 continuous features.} The Musk Anomaly Detection dataset was created to classify six classes of molecular musk. {This dataset has 3062 samples with 166 features.}  These four datasets, 20, 15, 2.5 and 3.2 percent of the data are anomalous samples, respectively. Hence, in the test phase, after calculating the anomaly score, the aforementioned proportion of the data that has the highest anomaly score is classified as an anomaly. In order to assess the proposed model on these datasets, F1, recall, and precision criteria are used.
Two datasets, CIFAR-10 \cite{krizhevsky2009learning} and SVHN \cite{netzer2011reading}, are considered for the image datasets. {CIFAR-10 consists of 60,000 32x32 color images containing one of 10 object classes, with 6000 images per class. SVHN dataset includes nearly 100000 32x32 labeled real world images of house numbers taken from Google Street View. Pixels of these two datasets were scaled to be in range [-1,1].} Both of these datasets have ten classes, and, like the previous works, one class is considered as the normal class and the other nine classes as the abnormal class. The model is trained only on normal data manifold. {As a result, we should train the model specifically for each class.} The criterion used to evaluate the model on the image dataset is the area under the receiver operating curve (AUROC). For all the datasets that are used, 80\% of the data are used for training, and 20\%  are used for testing. Validation data is chosen from 25\%  of the training data. 

\subsection{Experiments on the Tabular Datasets}

Evaluation results of the proposed RCALAD model and other state-of-the-art models on tabular datasets (KDDCup99, arrhythmia, thyroid, and musk) are summarized in Table \ref{table1}. The structures used in the generator, discriminator, and encoder networks are all fully connected layers with nonlinear activation functions. It should be noted that, in this step, $N(0, I)$ distribution is used as $\sigma(\mathrm{x})$.
In comparison to existing models, the proposed model has a successful performance on the arrhythmia and musk datasets, as shown in Table \ref{table1}. Our model is also the best according to F1 criteria on the KDD dataset, but it takes second place on the thyroid dataset due to the exceptional performance of the IF model. The reason for this phenomenon can be attributed to the nature of the data in this dataset. Since there are various features in this dataset, only a few of them are informative; therefore, the results of classic models such as IF, which are based on feature selection, are better. An idea to improve the proposed model results  on the thyroid dataset is to use models such as IF in the preprocessing step to select more informative features for training the model.

\begin{table*}[!htb]
\caption{Output results of the proposed model in comparison with the basic models on the tabular data set.}
\label{table1}
\begin{center}
\scalebox{1}{
\begin{tabular}{c|ccc|ccc|ccc|ccc} 
\hline
{Model} & \multicolumn{3}{c|}{ KDDCup~ } & \multicolumn{3}{c|}{ Arhythmia}            & \multicolumn{3}{c|}{ Thyroid~ }                                    & \multicolumn{3}{c}{ Musk }                                          \\ 
                       & Prec.   & Recall  & F1 score      & Prec.   & ~ ~Recall & F1 score                 & Prec.                & Recall               & F1 score                 & Prec.                & ~ ~Recall            & F1 score                 \\ 
\hline
IF\cite{25liu2008isolation}                     & $92.1$ & $93.7$ & $92.9$    & $51.4$ & $54.6$   & $53.0$              & $\mathbf{7 0 . 1 }$ & $\mathbf{7 1 . 4 }$ & $\mathbf{7 0 . 2 }$ & $47.9$              & $47.7$              & $47.5$               \\
OC-SVM\cite{12scholkopf1999support}                 & $74.5$ & $85.2$ & $79.5$    & $53.9$ & $40.8$   & $45.1$              & $36.3$              & $42.3$              & $38.8$              & $-$                  & $-$                  & $-$                   \\
DSEBMr\cite{20zhai2016deep}                 & $85.1$ & $64.7$ & $73.2$    & $15.1$ & $15.1$   & $15.1$              & $4.0$               & $4.03$               & $4.0$               & $-$                  & $-$                  & $-$                   \\
DSEBMe\cite{20zhai2016deep}                 & $86.1$ & $64.4$ & $73.9$    & $46.6$ & $45.6$   & $46.0$              & $13.1$              & $13.1$              & $13.1$              & $-$                  & $-$                  & $-$                   \\
AnoGAN\cite{8schlegl2017unsupervised}                 & $87.8$ & $82.9$ & $88.6$    & $41.1$ & $43.7$   & $42.4$              & $44.1$              & $46.8$              & $45.4$              & $3.0$               & $3.1$               & $3.1$                \\
DAGMM\cite{26zong2018deep}                  & $92.9$ & $94.2$ & $93.6$    & $49.0$ & $50.7$   & $49.8$              & $47.6$              & $48.3$              & $47.8$              & $-$                  & $-$                  & $-$                   \\
ALAD\cite{4zenati2018adversarially}                   & $94.2$ & $\mathbf{95.7}$ & $95.0$    & $50.0$ & $53.1$   & $51.5$              & $22.9$              & $21.5$              & $22.2$              & $58.1$              & $59.0$              & $58.3$               \\
DSVDD\cite{13ruff2018deep}                  & $89.8$ & $94.9$ & $92.1$    & $35.3$ & $34.3$   & $34.7$              & $22.2$              & $23.6$              & $23.2$              & $-$                  & $-$                  & $-$                   \\
RCALAD                 & $\mathbf{95.3}$ & $95.6$ & $\mathbf{95.4}$    & $\mathbf{58.8}$ & $\mathbf{62.5}$   & $\mathbf{6 0 . 6 }$ & $53.7$              & $51.5$              & $52.6$              & $\mathbf{6 2 . 9 }$ & $\mathbf{6 3 . 3 }$ & $\mathbf{6 3 . 1 }$  \\ 
\hline
\end{tabular}
}
\end{center}
\end{table*}
\subsection{Experiments on the image Datasets}

In this section, the performance of the proposed model on CIFAR-10 and SVHN image data is scrutinized in two separate tables. All the experiment are done in one vs. all setting. {It means that we consider one class as normal and the nine others as anomalies, we train our model only on the normal class. For each dataset, we train a specific model for each normal class.  We report the results for each class separately, as well as the average performance for whole classes. }

\begin{figure}[!htb]
  \centering
  \includegraphics[width=0.8\columnwidth]{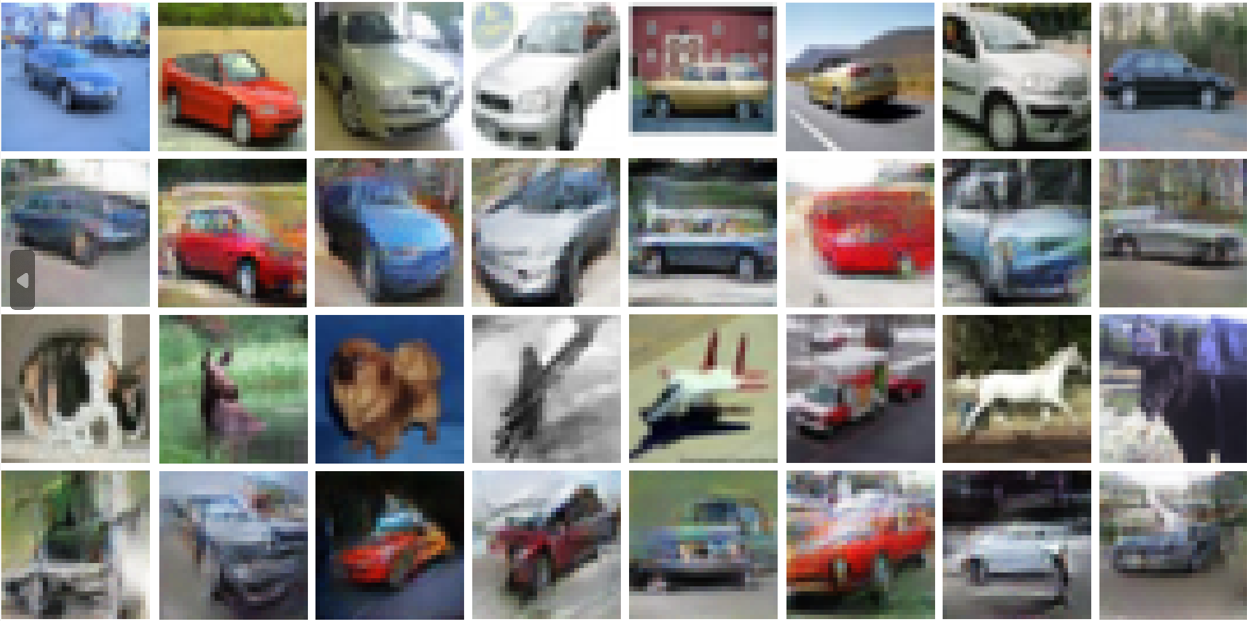}
   \caption{Reconstruction of normal and abnormal inputs on CIFAR-10 dataset. The first row is normal inputs and the second row is their reconstruction. The third row are anomalies and the fourth row is corresponding reconstruction for anomalies.}
   \label{fig5}
\end{figure}

\begin{figure}[!htb]
  \centering
  \includegraphics[width=0.8\columnwidth]{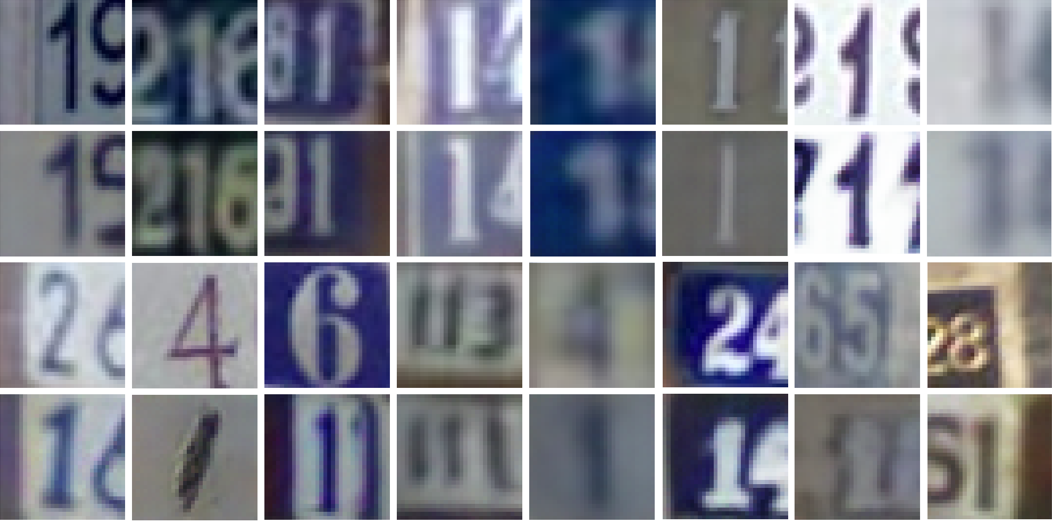}
   \caption{Reconstruction of normal and abnormal inputs on SVHN dataset. The first row is normal inputs and the second row is their reconstruction. The third row are anomalies and the fourth row is corresponding reconstruction for anomalies.}
   \label{fig6}
\end{figure}

As in Tables \ref{table2} and \ref{table3}, the proposed model has significantly improved the results on the CIFAR-10 dataset. The results show that the model performs best in the half of classes and gets competitive results on others. In addition to being superior in seven classes on SVHN dataset, the proposed model also performs the best in the average of all classes. Furthermore our model decreases class-wise variance on image datasets that leads to more reliable results. We provide some example of input images with their reconstruction in \ref{fig5} and \ref{fig6}. As you can see our model produces reconstructions in normal data manifold
even for anomalous samples. For example the proposed model can reconstruct a dog image as a car.{ In order to validate the superior performance of the RCALAD model, we also conducted the Wilcoxon  signed-rank test. In Appendix I, the P-values and statistics are reported.}

\begin{table*}[!ht]
\caption{Output results of the proposed model compared to the basic models on the CIFAR-10 dataset.}
\label{table2}
\begin{center}
\scalebox{1}{
\begin{tabular}{c|ccccccc} 
\hline
Normal   & DCAE\cite{27makhzani2015winner}                                  & DSEBM\cite{20zhai2016deep}          & DAGMM\cite{26zong2018deep}          & IF\cite{25liu2008isolation}             & AnoGAN\cite{8schlegl2017unsupervised}                                & ALAD\cite{4zenati2018adversarially}           & RCALAD                                 \\ 
\hline
Airplane & $59.1$                        & $41.4$ & $56.0$ & $60.1$ & $67.1$                        & $64.7$ & $\mathbf{68. 4}$  \\
car    & $\mathbf{5 7 . 4}$            & $57.1$ & $56.0$ & $50.8$ & $54.7$                        & $45.7$ & $57.2$                         \\
Bird     & $48.9$                        & $61.9$ & $53.8$ & $49.2$ & $52.9$                        & $67.0$ & $\mathbf{69 . 6}$             \\
Cat      & $58.4$                        & $50.1$ & $51.2$ & $55.1$ & $54.5$                        & $59.2$ & $\mathbf{6 7 . 2}$             \\
Deer     & $54.0$                        & $\mathbf{73.2}$ & $52.2$ & $49.8$ & $65.1$                        & $72.7$ & $71.9$             \\
Dog      & $6 2 . 2$          & $60.5$ & $49.3$ & $58.5$ & $60.3$                        & $52.8$ & $\mathbf{65.1}$                         \\
Frog     & $51.2$                        & $68.4$ & $64.9$ & $42.9$ & $58.5$                        & $69.5$ & $\mathbf{70 . 3}$             \\
Horse    & $58.6$                        & $53.3$ & $55.3$ & $55.1$ & $\mathbf{6 2 . 5}$ & $44.8$ & $59.6$                         \\
Ship     & $\mathbf{7 6 . 8} $          & $73.9$ & $51.9$ & $74.2$ & $75.8$                        & $73.4$ & $7 0 . 5$             \\
Truck    & $\mathbf{6 7 . 3}$ & $63.6$ & $54.2$ & $58.9$ & $66.5$                        & $43.2$ & $57.6$                         \\ 
\hline
Mean     & $59.4$                                & $60.3$         & $54.4$         & $55.5$         & $61.8$                                & $59.3$         & $\mathbf{6 5 . 7}$                     \\
\hline
\end{tabular}}
\end{center}
\end{table*}

\begin{table*}[!ht]
\caption{Output results of the proposed model compared to the basic models on the SVHN dataset.}
\label{table3}
\begin{center}
\scalebox{1}{
\begin{tabular}{c|ccccccc} 
\hline
Normal & OCSVM\cite{12scholkopf1999support}          & DSEBMr\cite{20zhai2016deep}         & DSEBMe\cite{20zhai2016deep}         & IF\cite{25liu2008isolation}             & AnoGAN\cite{8schlegl2017unsupervised}         & ALAD\cite{4zenati2018adversarially}                                  & RCALAD                              \\ 
\hline
0      & $52.0$ & $56.1$ & $53.4$ & $53.0$ & $57.3$ & $58.7$                        & $\mathbf{6 0 . 4}$          \\
1      & $48.6$ & $52.3$ & $52.1$ & $51.2$ & $57.0$ & $\mathbf{6 2 . 8}$ & $59.2$                      \\
2      & $49.7$ & $51.9$ & $51.8$ & $52.3$ & $53.1$ & $\mathbf{55 . 2}$    & $54.9$                      \\
3      & $50.9$ & $51.8$ & $51.7$ & $52.2$ & $52.6$ & $53.8$                        & $\mathbf{55.8}$  \\
4      & $48.4$ & $52.5$ & $52.4$ & $49.1$ & $53.9$ & $58.0$                        & $\mathbf{58.5}$    \\
5      & $51.1$ & $52.4$ & $52.3$ & $52.4$ & $52.8$ & $56.1$                        & $\mathbf{56.2}$    \\
6      & $50.1$ & $52.1$ & $52.2$ & $51.8$ & $53.2$ & $57.4$                        & $\mathbf{59.4}$    \\
7      & $49.6$ & $53.4$ & $55.3$ & $52.0$ & $55.0$ & $\mathbf{58.8}$      & $58.0$                      \\
8      & $45.0$ & $51.9$ & $52.5$ & $52.3$ & $52.2$ & $5 5 . 2$            & $\mathbf{56.1}$    \\
9      & $52.5$ & $55.8$ & $52.7$ & $53.7$ & $53.1$ & $57.3$                        & $\mathbf{5 8 . 3}$          \\ 
\hline
Mean   & $50.2$         & $52.9$         & $52.4$         & $51.6$         & $54.0$         & $57.3$                                & $57.7$                              \\
\hline
\end{tabular}}
\end{center}
\end{table*}
\subsection{Ablation Studies}

In this section, we examine the effectiveness of each component added to the basic model on both kinds of datasets. In these experiments, the average results of the model are repeated in the presence and absence of the discriminator $D_{xxzz}$ and the supplementary distribution $\sigma(\mathrm{x})$. {Through these experiments, we want to find out how each proposed part affects the final results.}

According to Tables \ref{table4} and \ref{table5}, the proposed RCALAD model achieves the best result in the presence of both parts. In scrutinizing the role of the $D_{xxzz}$ discriminator, this discriminator has improved the accuracy on the CIFAR-10 dataset to an optimal level but has not made significant improvement on the SVHN dataset. In terms of the role of the $\sigma(\mathrm{x})$ distribution, it performed well on the CIFAR-10 dataset and enhanced the AUROC criterion. However, when applied to the SVHN dataset, it reduced the AUROC criteria by a tiny amount when compared to the base model. Still, its inclusion in the final model resulted in the extraction of new information and a more comprehensive view. To sumup, when both $D_{xxzz}$ and $\sigma(\mathrm{x})$ are present in training phase, we can achieve the best results.

\begin{table*}[!ht]
\centering
\caption{Effects of the various proposed sections in improving the results of the tabular datasets.}
\label{table4}
\scalebox{0.9}{
\begin{tabular}{ l|c|c|c }
\multicolumn{1}{c|}{Model}   & Precision                                                                                  & Recall                                                                                                                               & F1 score                                                                                                \\ 
\hline
\multicolumn{4}{c}{KDDCup99}                                                                                                                                                                                                                                                                                                                                                  \\ 
\hline
Baseline(ALAD)              & $94.4$                                                              & $\mathbf{95.7} $                                           & $95.0$                                                                          \\
Baseline + $D_{xxzz}$(CALAD)     & $\mathbf{95.9}$ & $95.7$                                                                                                       & \begin{tabular}[c]{@{}c@{}}$\mathbf{95.8}$\\\end{tabular}  \\
Baseline + $\sigma(\mathrm{x})$(RALAD)           & $94.3$                                                             & $95.5$                                                                                                       & $94.9$                                                                          \\
Baseline + $D_{xxzz}$ + $\sigma(\mathrm{x})$(RCALAD) & $95.3$                                                             & $95.6$                                                                                                       & $95.4$                                                                          \\ 
\hline
\multicolumn{4}{c}{~~~~~~ Arrhythmia}                                                                                                                                                                                                                                                                                                                                      \\ 
\hline
Baseline(ALAD)              & $50.0$                                                             & $53.1$                                                                                                       & $51.5$                                                                          \\
Baseline + $D_{xxzz}$(CALAD)     & $57.4$                                                             & $60.5$                                                                                                       & $57.5$                                                                          \\
Baseline + $\sigma(\mathrm{x})$(RALAD)           & $54.6$                                                             & $56.5$                                                                                                       & $55.5$                                                                          \\
Baseline + $D_{xxzz}$ + $\sigma(\mathrm{x})$(RCALAD) &  $\mathbf{58.8}$                                & $\mathbf{62.5}$                                             & $\mathbf{60.6}$                                          \\ 
\hline
\multicolumn{4}{c}{Thyroid}                                                                                                                                                                                                                                                                                                                                                \\ 
\hline
Baseline(ALAD)              & $22.9$                                                             & $21.5$                                                                                                       & $22.2$                                                                          \\
Baseline + $D_{xxzz}$(CALAD)     & $52.9$                                                             & \begin{tabular}[c]{@{}c@{}}$\mathbf{51.8} $\\\end{tabular} & $52.3$                                                                          \\
Baseline + $\sigma(\mathrm{x})$(RALAD)           & $43.1$                                                             & $45.7$                                                                                                       & $44.3$                                                                          \\
Baseline + $D_{xxzz}$ + $\sigma(\mathrm{x})$(RCALAD) & $\mathbf{53.7}$ & $51.5$                                                                                                       & $\mathbf{52.6}$              \\ 
\hline
\multicolumn{4}{c}{Musk}                                                                                                                                                                                                                                                                                                                                                   \\ 
\hline
Baseline(ALAD)              & $50.0$                                                             & $53.1$                                                                                                       & $51.5$                                                                          \\
Baseline + $D_{xxzz}$(CALAD)     & $57.4$                                                             & $60.5$                                                                                                       & $57.5$                                                                          \\
Baseline + $\sigma(\mathrm{x})$(RALAD)           & $54.6$                                                             & $56.5$                                                                                                       & $55.5$                                                                          \\
Baseline + $D_{xxzz}$ + $\sigma(\mathrm{x})$(RCALAD) & $\mathbf{62.9}$ & $\mathbf{63 .3}$ & $\mathbf{63.1}$             
\end{tabular}}
\end{table*}

\begin{table*}[!htb]
\caption{Effects of the various proposed sections in improving the results of the image datasets.}
\label{table5}
\begin{center}
\scalebox{1}{
\begin{tabular}{l|c}

\multicolumn{1}{c|}{Model}& \multicolumn{1}{c}{AUROC}  \\ 
\hline\multicolumn{2}{c}{CIFAR-10}\\ \hline
Baseline(ALAD)                        & $59.3$                                                                 \\
Baseline + $D_{xxzz}$(CALAD)            & $63.4$                                                                 \\
Baseline + $\sigma(\mathrm{x})$(RALAD)                     & $64.2 $                                                                 \\
Baseline + $D_{xxzz}$ + $\sigma(\mathrm{x})$(RCALAD)  & $\mathbf{65.7}$                                                                 \\ 
\hline
\multicolumn{2}{c}{SVHN}                                                                                                   \\ 
\hline
Baseline(ALAD)                        & $57.3$                                                                 \\
Baseline + $D_{xxzz}$(CALAD)            & $57.6$                                                                 \\
Baseline + $\sigma(\mathrm{x})$(RALAD)                     & $56.8$                                                                 \\
Baseline + $D_{xxzz}$ + $\sigma(\mathrm{x})$(RCALAD)        & $\mathbf{57.7}$                  

\end{tabular}}
\end{center}
\end{table*}

\subsection{Evaluating the Sufficiency of the $D_{xxzz}$}
By adding the $D_{xxzz}$ discriminator, is there a need for $D_{xx}$ and $D_{zz}$ discriminators or not? To answer this question correctly, we performed some experiments, whose results are summarized in Table \ref{table6}. In fact, in this section, in addition to the above-mentioned question, the result of adding $D_{xxzz}$ discriminator in basic models such as ALI and ALICE are investigated.

In Table \ref{table6} we examine the effects of all discriminators on different models. According to this table and as expected from the theoretical results, adding the $D_{xxzz}$ discriminator to the general frameworks had the highest efficiency. As a result, eliminating $D_{xx}$ has less of an impact on the model because some of the information it collects is covered by the $D_{xxzz}$ discriminator. However, it is apparent that removing $D_{zz}$ , which assesses the similarity of $z$ and its reconstruction in an independent cycle, decreases the accuracy. As can be seen, employing these three discriminators is seen to be the most effective, as the $D_{xxzz}$ discriminator alone does not cover all aspects. It is worth noting that this remark applies to both image and tabular datasets.
\begin{table*}[!htb]
\caption{Assessing the performance of the model in the presence or absence of each of the discriminators.}
\label{table6}
\begin{center}
\scalebox{0.9}{
\begin{tabular}{l|ccc|ccc}
\multicolumn{1}{c|}{model}          & $D_{zz}$ & $D_{xx}$ & $D_{xxzz}$ & Precision                                                                              & Recall                                                                   & F1 score                                                                      \\ 
\hline
\multicolumn{7}{c}{KDDCup99}                                                                \\ 
\hline
ALAD                                & yes & yes & no    & $94.2$                                                     & $\mathbf{95.7}$       & $95.0$                                           \\
ALI
+ $D_{xxzz}$                         & no  & no  & yes   & $93.8$                                                     & $95.1$  & $94.4$                                           \\
ALI + $D_{zz}$+ $D_{xxzz}$ & yes & no  & yes   & $94.6$                                                     & $95.5$                                           & $95.0$                                           \\
ALICE + $D_{xxzz}$     & no  & yes & yes   & $94.1$                                                     & \begin{tabular}[c]{@{}c@{}}$94.5$\end{tabular} & \begin{tabular}[c]{@{}c@{}} $94.7$ \end{tabular}  \\
CALAD                               & yes & yes & yes   & $\mathbf{95.9}$ & $95.7$                                           & $\mathbf{95.8}$       \\
RCALAD                              & yes & yes & yes   & $95.3$                                                     & $95.6$                                           & $95.4$                                           \\ 
\hline
\multicolumn{7}{c}{Arrhythmia}                                                                                                                                                                                                                                                                     \\ 
\hline
ALAD                                & yes & yes & no    & $50.0$                                                     & $53.1$  & $51.5$                                           \\
ALI
+ $D_{xxzz}$                         & no  & no  & yes   & $52.2$                                                     & $52.9$   & $52.5$                                           \\
ALI+ $D_{zz}$ + $D_{xxzz}$ & yes & no  & yes   & $57.1$                                                     & $58.2$                                           & $57.6$                                           \\
ALICE + $D_{xxzz}$     & no  & yes & yes   & $54.3$                                                     & $56.1$                                           & $55.1$                                           \\
CALAD                               & yes & yes & yes   & $57.4$                                                     & $60.5$     & $57.5$                                           \\
RCALAD                              & yes & yes & yes   & $\mathbf{58.8}$                                 & $\mathbf{62.5}$                       & $\mathbf{60.6}$                      

\end{tabular}}
\end{center}
\end{table*}

\subsection{Scores Evaluation}
In this section, the proposed anomaly scores are evaluated and compared with the anomaly scores presented in previous work \cite{4zenati2018adversarially}. As shown in Table \ref{table7}, on tabular data, the raw output of the $D_{xxzz}$ discriminator $A_{all}$ outperforms other anomaly scores. The performance of the feature-based score $A_{fm}$ on image data is remarkable, as can be seen in Table \ref{table8}. This difference in the performance of the introduced scores can be attributed to the difference in the number of features in these two types of datasets. Given the fact that the number of features on tabular data is less than those on image data, the outputs of all discriminators are enough to detect anomalous data. However, in the image datasets, the output of the next-to-last layer contains more information to distinguish between normal and abnormal data. In this way, the $A_{fm}$ score excelled on the image datasets, while $A_{all}$ performed well on tabular datasets.
\begin{table*}
\caption{Comparing the performance of the proposed anomaly scores with other scores on tabular data}
\label{table7}
\begin{center}
\scalebox{1}{
\begin{tabular}{l|c|c|c}
\multicolumn{1}{c|}{Score} & Precision & Recall & F1 score \\ 
\hline
\multicolumn{4}{c}{KDDCup99} \\ 
\hline
$A_{L_1}$ & $90.8$ & $91.0$ & $90.9$ \\
$A_{L_2}$ & $90.1$ & $90.0$ & $90.0$ \\
$A_{Logits}$ & $91.6$ & $91.6$ & $91.6$ \\
$A_{Features}$ & $91.2$ & $91.7$ & $91.1$ \\
$A_{fm}$ & $\mathbf{93.2}$ & $\mathbf{93.7}$ & $\mathbf{93.0}$ \\
$A_{all}$ & $92.3$ & $90.0$ & $92.1$ \\ 
\hline
\multicolumn{4}{c}{Arrhythmia} \\ 
\hline
$A_{L_1}$ & $35.2$ & $37.5$ & $36.4$ \\
$A_{L_2}$ & $35.2$ & $37.5$ & $36.4$ \\
$A_{Logits}$ & $55.8$ & $59.3$ & $57.6$ \\
$A_{Features}$ & $23.2$ & $25.0$ & $24.2$ \\
$A_{fm}$ & $44.1$ & $46.8$ & $45.4$ \\
$A_{all}$ & $\mathbf{61.7}$ & $\mathbf{65.6}$ & $\mathbf{63.7}$ \\ 
\hline
\multicolumn{4}{c}{Thyroid} \\ 
\hline
$A_{L_1}$ & $49.8$ & $49.0$ & $49.9$ \\
$A_{L_2}$ & $50.1$ & $50.0$ & $50.0$ \\
$A_{Logits}$ & $49.6$ & $49.7$ & $49.7$ \\
$A_{features}$ & $51.2$ & $51.7$ & $51.5$ \\
$A_{fm}$ & $52.2$ & $51.2$ & $51.7$ \\
$A_{all}$ & $\mathbf{53.7}$ & $\mathbf{51.5}$ & $\mathbf{52.6}$ \\ 
\hline
\multicolumn{4}{c}{Musk} \\ 
\hline
$A_{L_1}$ & $59.7$ & $59.3$ & $59.5$ \\
$A_{L_2}$ & $60.0$ & $60.1$ & $60.1$ \\
$A_{Logits}$ & $58.6$ & $58.9$ & $58.8$ \\
$A_{Features}$ & $58.2$ & $58.8$ & $58.8$ \\
$A_{fm}$ & $61.1$ & $61.8$ & $61.4$ \\
$A_{all}$ & $\mathbf{62.9}$ & $\mathbf{63.3}$ & $\mathbf{63.1}$    
\end{tabular}}
\end{center}
\end{table*}

\begin{table*}[!h]
\caption{Comparing the performance of the proposed anomaly scores with other scores on image datasets.}
\label{table8}
\begin{center}
\scalebox{1.0}{
\begin{tabular}{l|c}
\multicolumn{1}{c|}{Anomaly Score} & AUROC     
                          \\ 
\hline
\multicolumn{2}{c}{CIFAR-10}                                                                                                             \\ 
\hline
$A_{L_1}$                             & $63.4$                                                                \\
$A_{L_2}$                            & $63.2$                                                                \\
$A_{Logits}$                &  $62.9$                                                                \\
$A_{Features}$              & $63.1$                                                                \\
$A_{fm}$                    & $\mathbf{65. 7}$
\\
$A_{all}$                    & $64.7$ 
\\ 
\hline
\multicolumn{2}{c}{SVHN}                                                                                                                \\ 
\hline
$A_{L_1}$                            & $57.7$                                                                \\
$A_{L_2}$                            & $56.3$                                                                                 \\
$A_{Logits}$                & $53.6$                                                                \\
$A_{Features}$              & $57.6$                                                                \\
$A_{fm}$                    & $\mathbf{57.7}$  \\
$A_{all}$                   & $57.6$                                        

\end{tabular}}
\end{center}
\end{table*}

\subsection{Choosing appropriate supplementary distribution}
In this section, we examine different probability distributions as $\sigma(\mathrm{x})$ in order to find the best option and its effect on the performance of the proposed model. We consider $N(0,I)$, $N(0,2I)$, and $U(-1,+1)$ as supplementary distributions. As we can see in table \ref{table9} $N(0,I)$ serves the best results.

\begin{table*}[!h]
\caption{Comparing the performance of different supplementary distribution on the RCALAD model.}
\label{table9}
\begin{center}
\scalebox{1}{
\begin{tabular}{c|ccc|ccc}
\hline & \multicolumn{3}{|c|}{ KDDCUP } & \multicolumn{3}{c}{ Arrhythmia } \\
                $\sigma(\mathrm{x})$ & Prec. & Recall & $F_1$ & Prec. & Recall & $F_1$ \\
                \hline $\mathcal{N}(\mathbf{0}, \boldsymbol{I})$ & $\mathbf{0 . 6 2 9}$ & $\mathbf{0 . 6 3 3}$ & $\mathbf{0 . 6 3 1}$ & $\mathbf{0 . 5 8 8}$ & 0.625 & 0.606 \\
                $\mathcal{N}(\mathbf{0}, 2 \boldsymbol{I})$ & 0.626 & $\mathbf{0 . 6 3 3}$ & 0.629 & 0.580 & 0.629 & 0.603 \\
                $\mathcal{U}(\mathbf{- 1}, \mathbf{1})$ & 0.608 & 0.604 & 0.606 & 0.584 & $\mathbf{0 . 6 3 3}$ & $\mathbf{0 . 6 0 7}$ \\
                \hline                                       

\end{tabular}}
\end{center}
\end{table*}
{
\subsection{Statistical Test}
    The Wilcoxon Rank Test \cite{Wilcoxon1945}, also known as the Mann Whitney U Test, is a non-parametric statistical test used to compare two independent samples. It is often described as the non-parametric version of the two-sample t-test. The test only makes the first two assumptions of independence and equal variance and does not assume that the data have a known distribution. The test is used to determine if there is a significant difference between the two groups being compared.
    \\ The research hypothesis is that the proposed method would have a higher accuracy than the previous methods. The null hypothesis is that there would be no difference in accuracy between the proposed method and the previous methods. As can be seen in Table 1-3, the highest accuracy after the proposed model belongs to the ALAD model, and for this reason, this statistical test has been performed with the results of this model. The result of this test is reported in Table \ref{table10}.
                    \\\begin{table*}[!ht]
                    \begin{center}
                    \caption{Result of Wilcoxom Rank Test.}
                    \label{table10}
                     \scalebox{1}{
                    \begin{tabular}{c|c|c|c} 
                        Dataset & \#run & statistics & P-value  \\
                        \hline Arrhythmia & 10 & 0.0 & 0.00195  \\
                        KDD99 & 10 & 7.0 & 0.03710 \\
                        SVHN & 30 & 157.0 & 0.12413 \\
                        CIFAR10 & 30 & 56.0 & 0.00011
                    \end{tabular}
                    }
                    \end{center}
                    \end{table*}
    \\The results of the Wilcoxon signed-rank test indicate that the proposed method had a significantly better performance than the previous methods in three over four datasets. As mentioned in section 5.1, Dataset SVHN pertains to real images of the house number. This dataset proved to be particularly challenging due to the presence of noise.
    To further investigate the performance of the proposed method on the SVHN dataset, we conduct a follow-up experiment consisting of 10 tests per class, as opposed to 3 tests described in Section 5. The results of this extended experiment yield a test statistic value of 1898.0 and a corresponding p-value of 0.03109, providing evidence of the superior performance of the proposed method on the challenging SVHN dataset.
\section{Conclusion}
 {This research presents a unique and novel solution for high variance rate problem in anomaly detection tasks through the use of generative adversarial neural networks (GANs). The proposed framework utilizes a generator and an encoder to learn the mapping between the input data space and latent space and incorporates two discriminators, $D_{xx}$ and $D_{zz}$, to improve the stability of the training process and ensure a consistent cycle. To further enhance the information utilization of the cycle, a new discriminator, $D_{xxzz}$, is introduced. Experiments reveal that considering cycle consistency for input data for the latent space variable $z$ as one complete joint cycle provides additional information for the model that improves the results. The proposed model also incorporates a supplementary distribution $\sigma(\mathrm{x})$ to influence the network output to align with the normal data distribution, leading to even more precise anomaly detection. The supplementary distribution helps the model map unseen data in the normal data manifold, and as a result, an appropriate distance between the anomalous sample and its reconstruction will be achieved.} The proposed model outperforms existing models for both tabular and image datasets in terms of anomaly detection and reducing class-wise variance in image datasets.

{
\section{Future Work}
The RCALAD model has yielded promising results in the field of anomaly detection. However, like other GAN-based models, it is susceptible to robustness challenges. To address these issues, existing techniques, as outlined in \cite{28salimans2016improved,29chalapathy2017robust} can be employed.
\\In the present study, the anomaly ratio parameter was an input during test time. A potential avenue for future research would be to explore methods that can effectively determine the optimal value for this parameter without requiring it to be supplied as input.
\\Further, another potential area for future direction could be learning importance coefficients for each discriminator based on the type of data being analyzed (e.g., tabular or imagery data) and incorporating these coefficients into the cost function for calculating anomaly scores.
}

\bibliographystyle{IEEEtran}
\bibliography{mybibfile}

\end{document}